\documentclass[conference]{IEEEtran}
\pdfoutput=1 

\usepackage[acronym]{glossaries}
\usepackage{multirow}
\usepackage{amssymb}
\usepackage{mathtools}

\DeclarePairedDelimiter\floor{\lfloor}{\rfloor}
\usepackage{grffile}
\usepackage{booktabs} 
\usepackage{subfig}
\usepackage{flushend}
\usepackage{times}
\usepackage{textcomp}
\usepackage{tabulary}
\usepackage{color}

\usepackage[font={small,it}]{caption}
\begin{document}
%
\title{Parallel Multi Channel Convolution \\using General Matrix Multiplication}

\author{\IEEEauthorblockN{Aravind Vasudevan}
\IEEEauthorblockA{School of Computer Science\\and Statistics\\Trinity College Dublin}
\and
\IEEEauthorblockN{Andrew Anderson}
\IEEEauthorblockA{School of Computer Science\\and Statistics\\Trinity College Dublin}
\and
\IEEEauthorblockN{David Gregg}
\IEEEauthorblockA{School of Computer Science\\and Statistics\\Trinity College Dublin}}
\newcommand{\aravind}[1]{*** AV : {\color{red} #1 }***}
\newcommand{\andrew}[1]{*** AA : {\color{blue} #1 }*** }
\newcommand{\david}[1]{*** DG : {\color{green} #1 }***}
\newcommand{\needsmassaging}[1]{Needs some massaging --- {\color{blue} #1 }}

\newcommand*\concat{\mathbin{\|}}

\newacronym{dnn}{DNN}{Deep Neural Network}
\newacronym{cnn}{CNN}{Convolutional Neural Network}

\newacronym{scsk}{SCSK}{Single Channel Single Kernel}
\newacronym{mcsk}{MCSK}{Multiple Channel Single Kernel}
\newacronym{mcmk}{MCMK}{Multiple Channel Multiple Kernel}

\newacronym{gemv}{GEMV}{General Matrix Vector multiplication}
\newacronym{gemm}{GEMM}{General Matrix Multiplication}

\newcommand{\ifm}{\mathcal{I}}			
\newcommand{\ofm}{\mathcal{O}}		 
\newcommand{\kernel}{\mathcal{K}}

\newcommand{\iw}{W}
\newcommand{\ih}{H}
\newcommand{\ic}{C}

\newcommand{\kw}{k}
\newcommand{\kh}{k}
\newcommand{\kc}{C}
\newcommand{\kn}{M}

\newcommand{\ow}{P}
\newcommand{\oh}{Q}
\newcommand{\oc}{\kn}

\newacronym{i2c}{$im2col$}{im2col}
\newacronym{i2r}{$im2row$}{im2row}

\newacronym{googlenet}{GoogLeNet}{GoogLeNet}
\newacronym{alexnet}{AlexNet}{AlexNet}
\newacronym{vgg}{VGG-16}{VGG-16}
\glsunset{googlenet}
\glsunset{alexnet}
\glsunset{vgg}

\newacronym{k2r}{$kn2row$}{Kernel to Row} 
\newacronym{k2c}{$kn2col$}{Kernel to Column} 

%
\maketitle

\glsunset{i2c}
\begin{abstract}
Convolutional neural networks (CNNs) have emerged as one of the most successful machine learning technologies for image and video processing. The most computationally-intensive parts of CNNs are the convolutional layers, which convolve multi-channel images with multiple kernels. A common approach to implementing convolutional layers is to expand the image into a column matrix (\gls{i2c}) and perform \gls{mcmk} convolution using an existing parallel \glsfirst{gemm} library. This \gls{i2c} conversion greatly increases the memory footprint of the input matrix and reduces data locality.

In this paper we propose a new approach to MCMK convolution that is based on \gls{gemm}, but not on \gls{i2c}. Our algorithm eliminates the need for data replication on the input thereby enabling us to apply the convolution kernels on the input images directly. We have implemented several variants of our algorithm on a CPU processor and an embedded ARM processor. On the CPU, our algorithm is faster than \gls{i2c} in most cases.

\end{abstract}

\IEEEpeerreviewmaketitle

\section{Introduction}
\label{sec:intr}
\glspl{cnn} are one of the most effective machine learning approaches
for a variety of important real world problems.  \glspl{cnn} require
very large amounts of computation for both training and
inference. \glspl{cnn} are constructed from networks of standard
components, such as \emph{convolution layers}, \emph{activation
  layers} and \emph{fully-connected layers}. In most successful
\glspl{cnn}, the great majority of computation is performed in the
convolutional layers.

\glspl{cnn} require a very large amount of computation, so it is important to
make best use of available hardware resources.  This hardware may take the form
of a standard CPU, or an accelerator such as a graphics processing unit (GPU),
digital-signal processor (DSP), or vector architecture.  However, making best
use of the hardware for computationally intensive problems often requires
careful tuning of code to make best use of the memory hierarchy, registers, and
vector parallelism. For example, processor/accelerator companies devote very
large effort to tuning the performance of standard operators such as those in
the the Basic Linear Algebra Subroutines (BLAS) \cite{Lawson:1979}.

When implementing \glspl{cnn} on a new accelerator or processor, it is
fortunately possible to exploit existing pre-tuned BLAS routines. In
particular, the BLAS general matrix multiplication (\glsfirst{gemm}) routine is
commonly used to implement DNN convolution. It is well-known that 2D
convolution can be implemented using matrix multiplication by converting one of
the input matrices to a Toeplitz matrix. This involves replicating image pixels
multiple times across different matrix columns. Once the Toeplitz matrix has
been constructed, convolution can be implemented using a highly-tuned
\gls{gemm} for the target architecture.


The \gls{i2c} approach has been highly successful in \gls{dnn} frameworks such
as Caffe, Theano and Torch \cite{Chetlur:2014}. However, a major downside of
\gls{i2c} is the space explosion caused by building the column matrix. For a
convolution with a 2D $k \times k$ kernel matrix, the column matrix is $k^2$
times larger than the original image.  Deep learning systems are often most
useful when deployed in the field, but the space required for the column matrix
may be far too large to fit in the memory of an embedded system. Even outside
of the embedded context, the increased memory requirement may stretch the
limits of on-chip local memories and caches, which may increase execution time
and memory traffic.



In this paper we propose a new approach to \gls{dnn} convolution that allows us
to exploit existing optimized routines for accelerators and processors, but
does not costly input transformation operations. We make a number of
contributions:


\begin{itemize}
  \item
   We formulate the problem to operate on a \textit{non-replicated}
   input image. This allows us to pose the problem as either one or a sequence of matrix multiplications.
\item We present an experimental evaluation of our approach
  on an embedded processor (ARM\textregistered ~Cortex\textregistered-A57), and
  a general purpose CPU (Intel\textregistered ~Core\texttrademark ~i5-4570) using highly-optimized
  parallel versions of \gls{gemm}.
\item Our new \gls{gemm}-based approaches perform better than \gls{i2c} in a great majority of the scenarios tested.
\end{itemize}

The remainder of this paper is organized as follows. Section~\ref{sec:back}
provides additional background and detail on the \gls{mcmk} convolution
operation which is central to deep neural networks. Section~\ref{sec:meth}
describes how convolution can be implemented with a column matrix. We also show
how our proposed approach retains the advantages of re-using \gls{gemm} for the
computationally-intensive tasks, but with improved data locality.
Section~\ref{sec:expe} presents an evaluation of a number of variants of our
approach. Finally, Section~\ref{sec:rela-work} describes related work.


%

\section{Background}
\label{sec:back}

\begin{figure}[t]
	\centering
	\includegraphics[width=\linewidth]{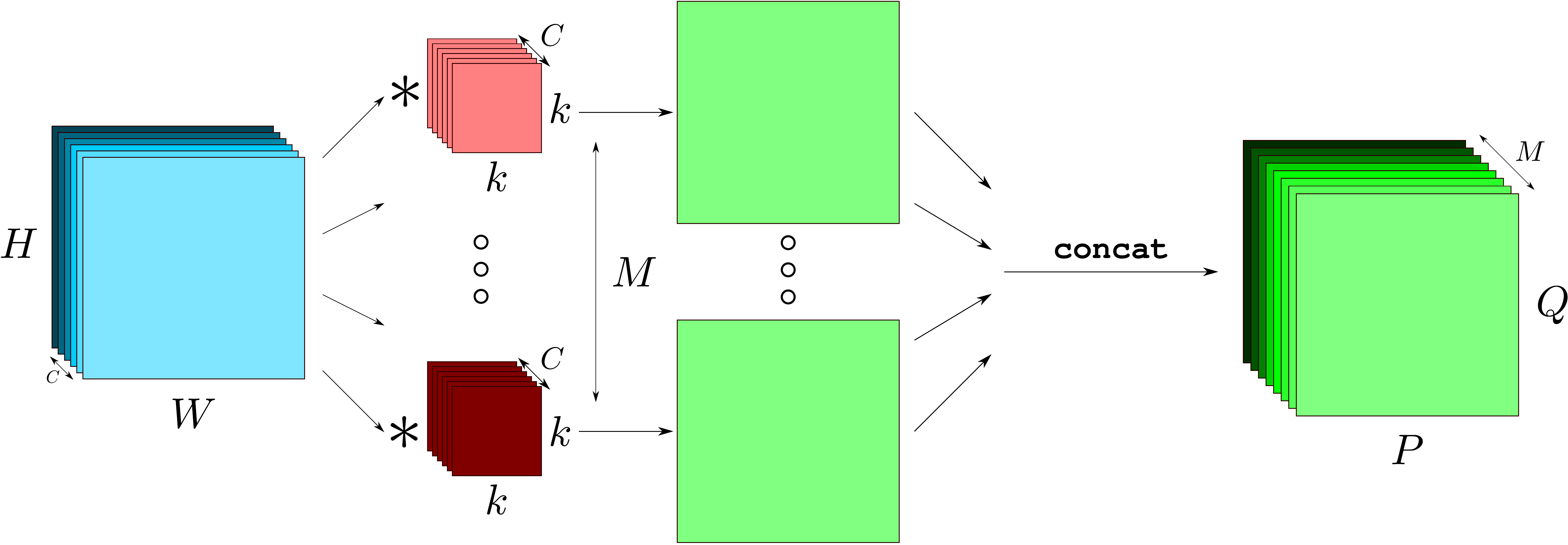}
	\caption{\glsfirst{mcmk} Convolution}
	\label{fig:mcmk}
\end{figure}


In this section, we give an overview of the \glsfirst{cnn}. Traditionally,
\glspl{cnn} are composed of a number of basic building blocks (often referred
to as \emph{layers}): \textbf{convolutional} layer, \textbf{pooling} layer
(average and max pooling), \textbf{activation} layer, \textbf{fully connected}
layer (FC) and \textbf{loss} layer.

\begin{figure*}[t]
	\centering
	\includegraphics[width=.6\linewidth]{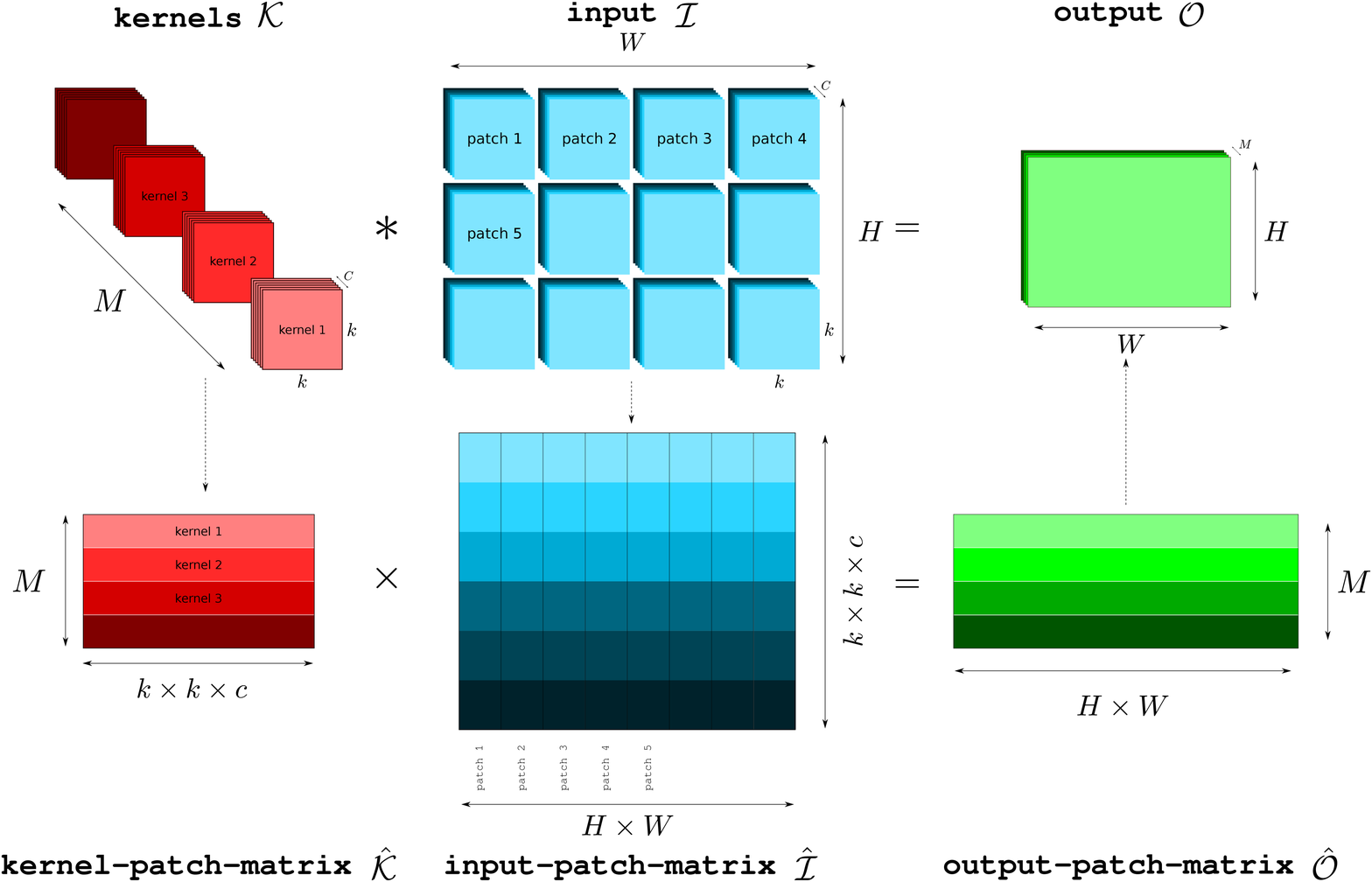}
	\caption{\gls{mcmk} using the \gls{i2c} method}
	\label{fig:im2col}
\end{figure*}
\subsection{Multi-channel convolution as sum of single channel convolutions}
\label{sec:mcmk}

In order to understand the core component of a \gls{cnn} -- the convolutional layer, we define dimensions of the inputs, the convolutional kernels (also known as feature maps) and the outputs. A convolutional layer (without batching) takes as input a 3D tensor (input -- $\ifm$) and a 4D tensor (kernel -- $\kernel$) and outputs a 3D tensor (output -- $\ofm$). At the core of any convolutional layer is a \gls{scsk}.

Given an input $\ifm \in \mathbb{R}^{\ih\times\iw}$ and a kernel $\kernel \in \mathbb{R}^{\kh\times\kw}$ an output element at position $(x,y)$ where $x\in[0,\oh)$ and $y\in[0,\ow)$, from the output $\ofm\in\mathbb{R}^{\oh\times\ow}$ is given by

\begin{multline}
\label{eq:scsk}
	conv2D_{x,y}(\ifm, \kernel) = \sum_{i=0}^{i=\kw-1} \sum_{j=0}^{j=\kh-1} \ifm( x-\floor{\frac{\kw}{2}}+i,\\ y-\floor{\frac{\kh}{2}}+j )\times\kernel( i, j )
\end{multline}

\noindent where $\iw$ and $\ih$ are the width and height of the input; $\kh$ is the size of the square kernel; $\ow$ and $\oh$ are the width and height of the output which is normally equal to $\iw$ and $\ih$ respectively\footnote{Note that  $\ow$ and $\oh$ are not equal to $\iw$ and $\ih$ if the convolution is ``strided''.
We do not consider strided convolutions in this paper as they account for only a small proportion of computation in most \glspl{cnn} 
}; and assuming the input is properly padded.



The \gls{mcsk} is then constructed using \gls{scsk}, by adding the result of 2D convolutions of the $\ic$ corresponding channels of the input $\ifm$ and the kernel $\kernel$. This is represented as

\begin{equation}
\label{eq:mcsk}
	\gls{mcsk}(\ifm_{\ic}, \kernel_{\ic}) = \sum_{c=0}^{c=\ic-1} conv2D(\ifm(c), \kernel(c))
\end{equation}

\noindent where $\ic$ is the number of channels in the input and kernel and $\ifm(c)$ and $\kernel(c)$ represent the $c$\textsuperscript{th} channel of the input and the kernel respectively. It is imperative to note that the output $\ofm\in\mathbb{R}^{\oh\times\ow}$ from Equation~\ref{eq:scsk} is a two-dimensional matrix and so is the output of \gls{mcsk} as shown in~\ref{eq:mcsk}. As discussed earlier, the convolution layer does \gls{mcmk} which can expressed as the concatenation of the resultant matrices from Equation~\ref{eq:mcsk} as shown in Figure~\ref{fig:mcmk} and is represented as

\begin{equation}
\label{eq:mcmk}
\resizebox{\linewidth}{!}
{
$\gls{mcmk}(\ifm_{\ic}, \kernel_{\ic}^{\kn}) = \gls{mcsk}(\ifm_{\ic}, \kernel_{\ic}^{0}) \concat \cdots \concat \gls{mcsk}(\ifm_{\ic}, \kernel_{\ic}^{\kn})$
}
\end{equation}

\noindent where $\kernel_{\ic}^{\kn}$ represents $\kn$ kernels with $\ic$ channels each and the $\concat$ operator denotes the concatenation of two channels.

\subsection{\gls{i2c}}
\label{sec:im2c}

The \gls{i2c} approach~\cite{chellapilla2006high,tsai2016performance,yanai2016efficient,gu2016opencl} has been well studied for transforming the \gls{mcmk} problem into a \gls{gemm} problem. Consider an input $\ifm \in \mathbb{R}^{\ih\times\iw\times\ic}$ and $\kn$ kernels $\kernel\in\mathbb{R}^{\kn\times\kh\times\kw\times\kc}$ as shown in Figure~\ref{fig:im2col}. From the input $\ifm$ we construct a new \emph{input-patch-matrix} $\hat{\ifm}$, by copying \emph{patches} out of the input and unrolling them into columns of this intermediate matrix. These patches are formed in the shape of the kernel (i.e. $\kh\times\kw\times\kc$) at every location in the input where the kernel is to be applied.


Once the input-patch-matrix $\hat{\ifm}$ is formed, we construct the kernel-patch-matrix $\hat{\kernel}$ by unrolling each of the $\kn$ kernels of the shape $\kh\times\kw\times\kc$ into a row of $\hat{\kernel}$ as shown in Figure~\ref{fig:im2col}. Note that this step can be avoided if the kernels are stored in this format to begin with (innermost dimension is the channel which forces the values along a channel to be contiguous). Then we simply perform a \gls{gemm} of $\hat{\kernel}$ and $\hat{\ifm}$ to get the output $\hat{\ofm}\in\mathbb{R}^{\ih\times\iw\times\kn}$ as shown in the figure.

It is easy to see from the above discussion how one could implement another method called \gls{i2r} wherein the local patches are unrolled into rows of the input-patch-matrix $\hat{\ifm}$ and the kernels are unrolled into columns of the kernel-patch-matrix $\hat{\kernel}$. We then perform a \gls{gemm} of $\hat{\ifm}$ and $\hat{\kernel}$ instead of $\hat{\kernel}\times\hat{\ifm}$ as in \gls{i2c}.

\vspace{.7cm}
\begin{figure}[h]
	\begin{minipage}{1\linewidth}
		\begin{footnotesize}
			\begin{verbatim}
			input[C][H][W];
			kernels[M][K][K][C];
			output[M][H][W];
			for h in 1 to H do
			 for w in 1 to W do
			  for o in 1 to M do
			   sum = 0;
			   for x in 1 to K do
			    for y in 1 to K do
			     for i in 1 to C do
			      sum += input[i][h+y][w+x] *kernels[o][x][y][i];
			   output[o][w][h] = sum;
			\end{verbatim}
		\end{footnotesize}
	\end{minipage}
	\caption{Simplified code for 2D multi-channel convolution with a single multi-channel input and multiple multi-channel convolution kernels. Note that special treatment of edge boundaries is not shown in this code.}
	\label{fig:covCode}
\end{figure}
\vspace{.7cm}

\begin{figure*}[t]
	\centering
	\includegraphics[width=.85\linewidth]{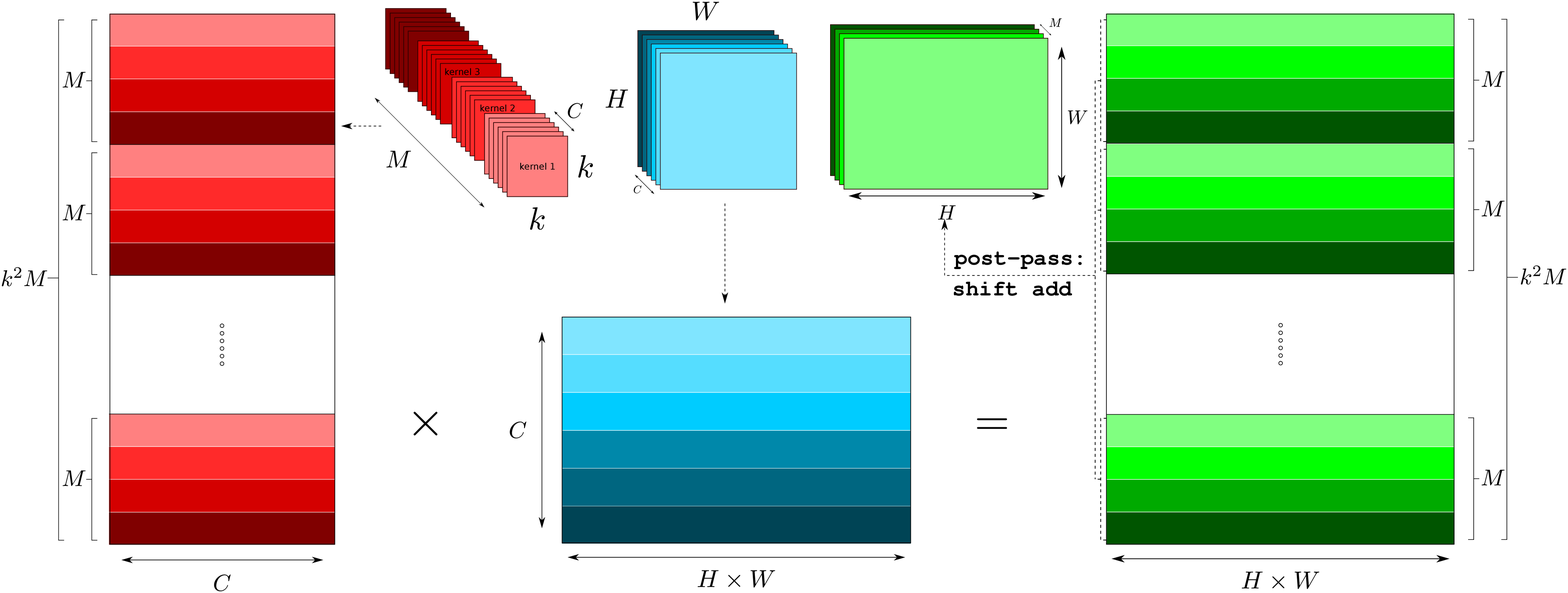}
	\caption{\gls{mcmk} using the ``\gls{k2r}'' method}
	\label{fig:k2r}
\end{figure*}


\section{A new approach}
\label{sec:meth}





A disadvantage of the \gls{i2c} approach is that it replicates input data to create the input-patch-matrix. For convolution with a $\kh \times \kw$ kernel, the input-patch-matrix matrix can be $\kh^2$ larger than the original input. A \gls{gemm}-based \gls{mcmk} algorithm that does not require data replication in the input could be useful for memory-limited embedded systems and might significantly improve data locality on any target architecture. In this section we present two \gls{gemm}-based \gls{mcmk} algorithms that eliminate data replication on the input, at the cost of some increase in the size of the output.


Figure~\ref{fig:covCode} shows a simplified loop nest for $\kh \times \kw$ convolution with $\kn$ kernels each with $\ic$ channels. A common operation in \glspl{cnn} such a \gls{googlenet}~\cite{Szegedy:2015} is convolution with a set of $1 \times 1$ convolutions. If we consider the code in figure \ref{fig:covCode} for the case where $\kh=1$, then the $x$ and $y$ loops collapse into a single iteration.

The resulting code is equivalent to 2D matrix multiplication of a $\kn \times \kc$ kernel times a $[\ic] \times [\ih \times \iw]$ input which results in a $[\kn] \times [\ih \times \iw]$ output. This output however is actually $\kn$ planes of $\ih\times\iw$ pixels which corresponds to an output of size $[\ih]\times[\iw]$ and $\kn$ channels. Let us call this correspondence of a $[\kn] \times [\ih \times \iw]$ matrix to an output matrix of size $[\ih]\times[\iw]$ and $\kn$ channels its \emph{multi-channel representation}, which we will use throughout the rest of this section. In other words, $1 \times 1$ \gls{mcmk} can be implemented by simply calling \gls{gemm} without data replication.

\subsection{\gls{k2r} and \gls{k2c}}
\label{sec:k2r}


Given that we can compute $1 \times 1$ \gls{mcmk} without data replication, how can we implement $\kh \times \kw$ \gls{mcmk}, for $\kh > 1$? We argue that a $\kh \times \kw$ convolution can be expressed as the sum of $\kh^2$ separate $1 \times 1$ convolutions. However the sum is not trivial to compute. Each $1 \times 1$ convolution yields a result matrix with dimensions $[\kn] \times [\ih \times \iw]$. We cannot simply add each of the resulting matrices pointwise, as each resultant matrix corresponds to a different kernel value in the $k\times k$ kernel. The addition of these matrices can then be resolved by offsetting every pixel in every channel of the \emph{multi-channel representation} of these matrices, vertically and/or horizontally (row and column offsets) by one or more positions before the addition.

For example, when computing a $3 \times 3$ convolution the result from computing the $1 \times 1$ \gls{mcmk} for the central point of the $3 \times 3$ kernel is perfectly aligned with the final sum matrix. On the other hand, the matrix that results from computing the $1 \times 1$ \gls{mcmk} for the upper left value of the $3 \times 3$ kernel must be offset up by one place and left by one place (in its \emph{multi-channel representation}) before being added to the final sum that computes the $3 \times 3$ \gls{mcmk}. Note that when intermediate results of $1 \times 1$ convolutions are offset, some values of the offsetted matrix fall outside the boundaries of the final result matrix. These out-of-bounds values are simply discarded when computing the sum of $1 \times 1$ convolutions.

It is possible to compute each of the $\kh^2$ separate $1 \times 1$ convolutions using a single matrix multiplication. We re-order the kernel matrix, so that the channel data is laid out contiguously, i.e. $\kn$ is the outer dimension and $\ic$ the inner. This data re-arrangement can be made statically ahead of time and used for all \gls{mcmk} invocations thereafter. Using a single call to \gls{gemm}, we multiply a $[\kh^2 \times \kn] \times [\ic]$ kernel matrix by a $[\ic] \times [\ih \times \iw]$ input matrix, resulting in a $[\kh^2 \times \kn] \times [\ih \times \iw]$ matrix. We perform a post pass of \texttt{\textbf{shift-add}} by summing each of the $\kn^2$ submatrices of size $\kn\times[\ih\times\iw]$ using appropriate offsetting in the \emph{multi-channel representation}. The result of this sum is a $[\kn] \times [\ih \times \iw]$ matrix, which is the output of our \gls{mcmk} algorithm. We refer to this as the \gls{k2r} algorithm.

If we swap the dimensions of the kernel matrix so that $\ic$ is not the innermost dimension and swap the input layout to make $\ic$ the innermost dimension, we get the \gls{k2c} algorithm. The \gls{gemm} call in this method would be to multiply an $[\ih \times \iw]\times[\ic]$ input matrix by a $[\ic]\times[\kh^2 \times \kn]$ kernel matrix, resulting in a $[\ih \times \iw]\times[\kh^2 \times \kn]$ matrix.

\section{Experiments and Results}
\label{sec:expe}

\glsunset{gemm}

\noindent
We evaluated the proposed \gls{mcmk} implementations on two general-purpose processors (one embedded, one desktop-class). The experimental platforms we used were ARM\textregistered ~Cortex\textregistered-A57 processor, which has 4 cores with a 128-bit wide SIMD unit, and the Intel\textregistered ~Core\texttrademark ~i5-4570, which has 4 cores and a 256-bit wide SIMD unit.

We used GCC version 7.1 to compile our code for the Intel and ARM CPUs. We used the latest stable version (0.2.19) of the high-performance OpenBLAS library to provide the \gls{gemm} operation on both ARM and Intel platforms.

We implemented a selection of \gls{mcmk} operations from three popular \gls{cnn} architectures: AlexNet~\cite{Krizhevsky:12}, VGG-16~\cite{Simonyan14c}, and \gls{googlenet}~\cite{Szegedy:2015}. In addition to our proposed \gls{gemm} based methods, we also implemented a direct convolution to provide some context for performance.

We experimented with several variants of direct convolution, including a version that is used in Caffe \cite{jia2014caffe}, and an optimized loop nest that appears in a recent book on optimizing code for the Intel Xeon Phi processor \cite{jeffers2016intel}. We found that the fastest direct method, on average, was actually the reference method: summation of single channel convolutions (Equation~\ref{eq:mcsk}).

We also benchmarked the convolution from Intel's MKL-DNN framework, which is the backend used by Intel Caffe. MKL-DNN supports AVX2 and AVX-512 processors, and incorporates a code generator which produces highly-optimized SIMD code for convolution.

We found that our \gls{gemm} based methods were often much faster than any direct method, and often outperform even the highly-optimized code produced by Intel's MKL-DNN.



\begin{figure*}[t]
	\centering
	\subfloat[VGG-16]{
		\centering
		\includegraphics[scale=0.58]{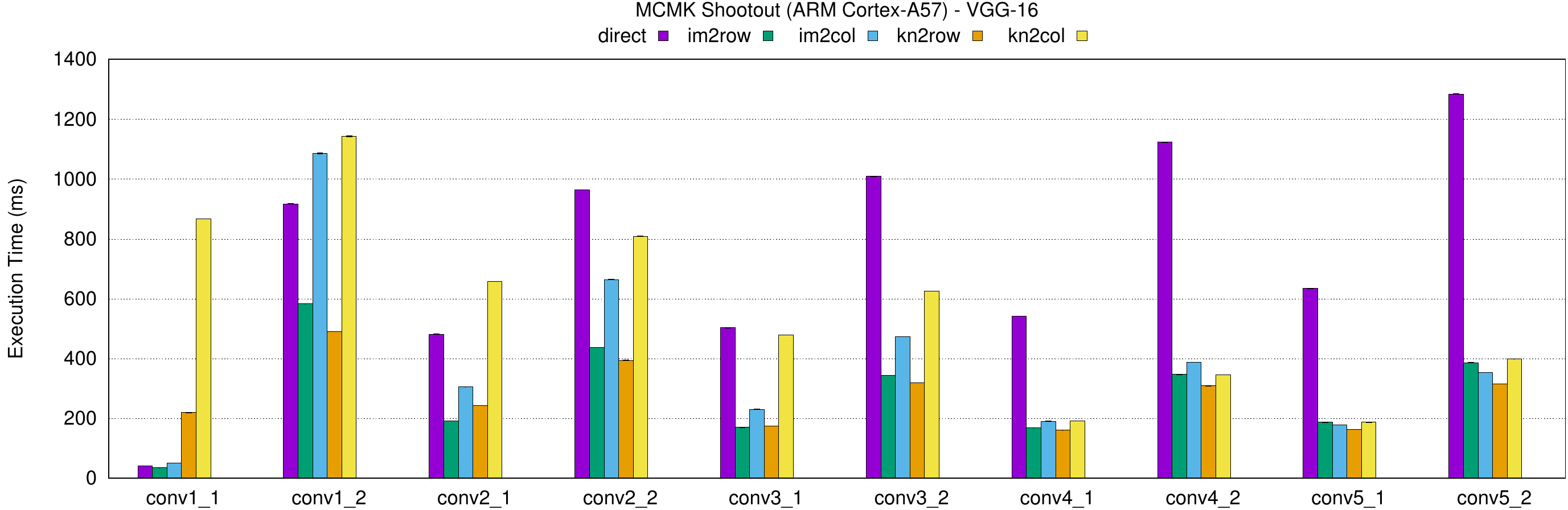}
		\label{fig:tx1-cpu-vgg}
	}\qquad
	\subfloat[Googlenet]{
		\centering
		\includegraphics[scale=0.52]{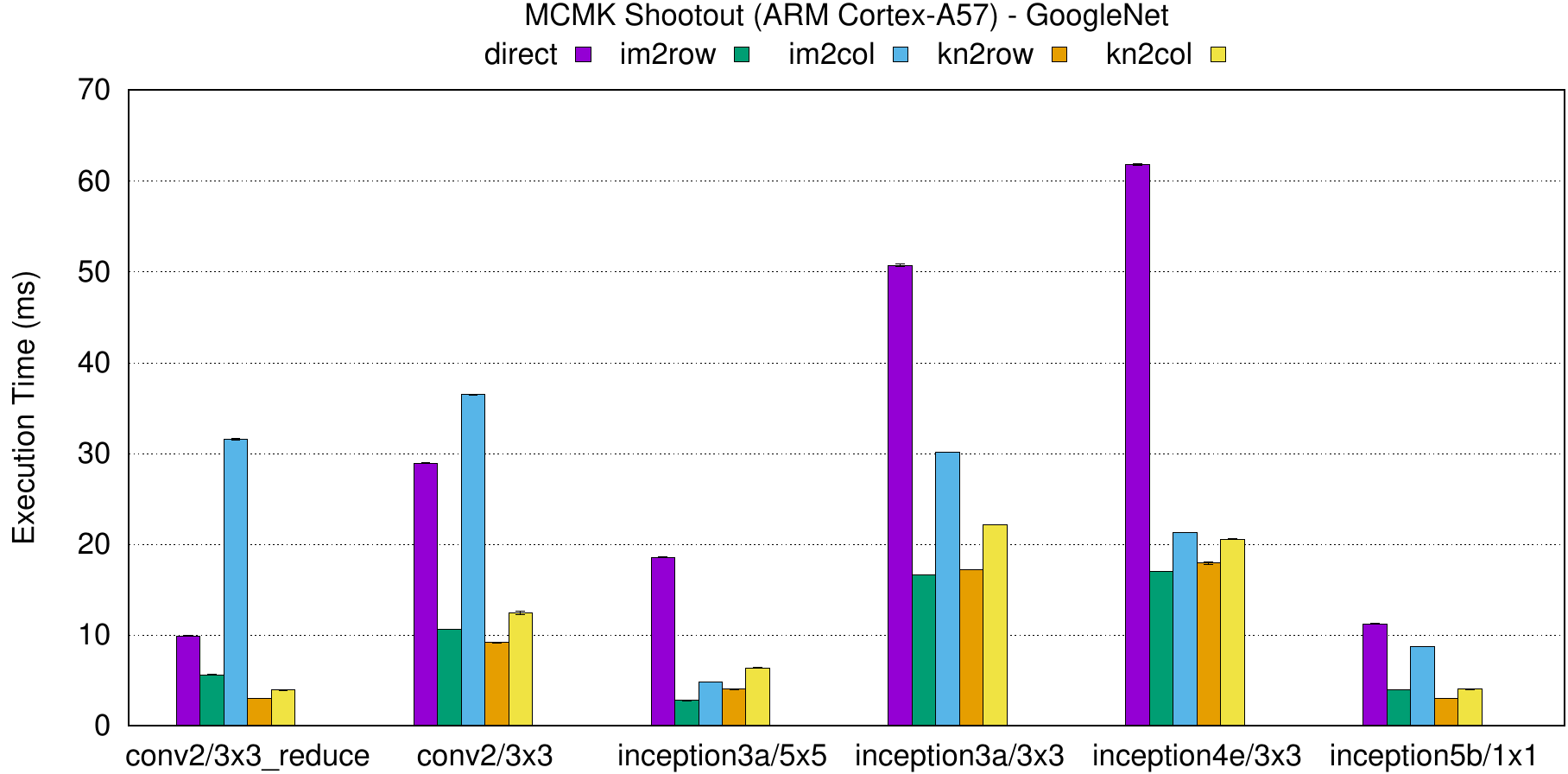}
		\label{fig:tx1-cpu-googlenet}
	}
	\subfloat[Alexnet]{
		\centering
		\includegraphics[scale=0.52]{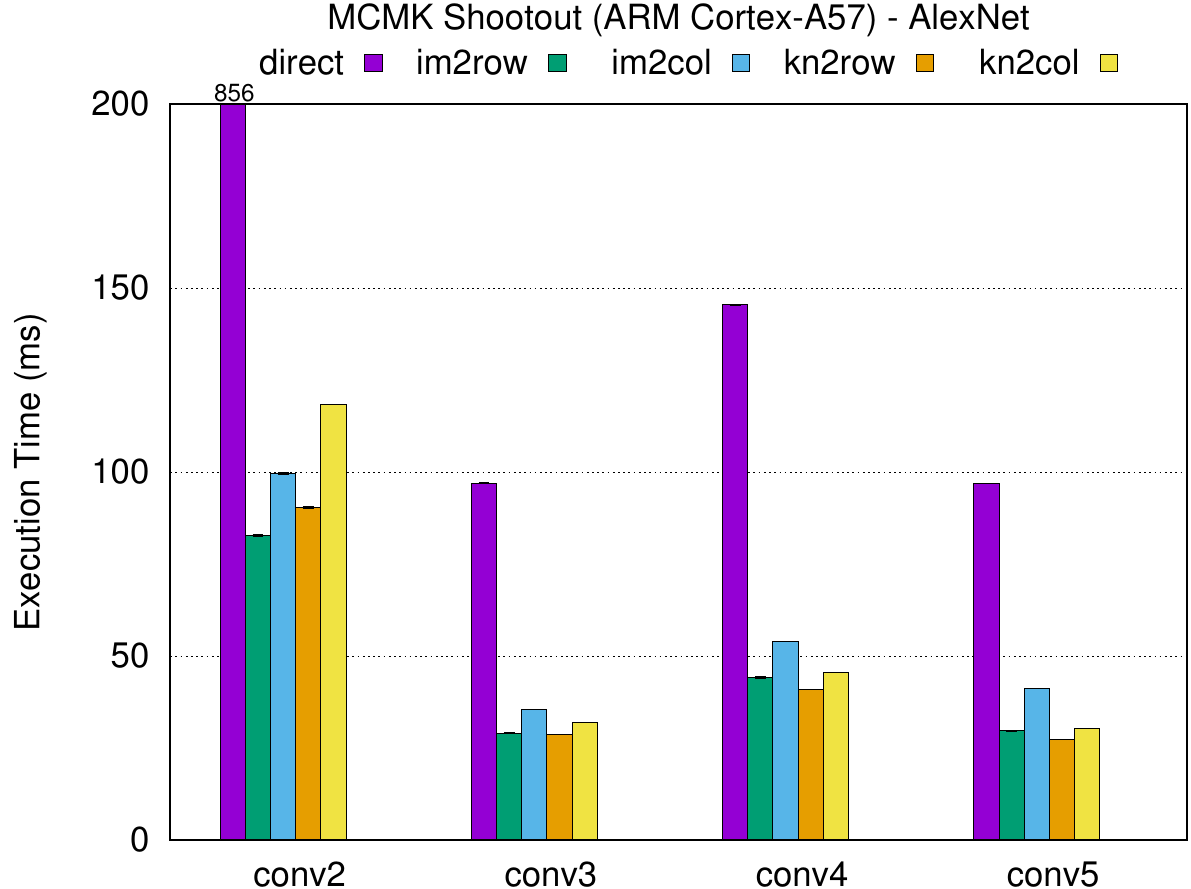}
		\label{fig:tx1-cpu-alexnet}
	}

	\caption{Execution time for selected layers of \gls{googlenet}, \gls{vgg} and \gls{alexnet} on the ARM\textregistered ~Cortex\textregistered-A57 CPU. \textbf{Lower is better}. }
	\label{fig:tx1-cpu-summary}
\end{figure*}

\begin{figure*}[h]
  \centering
    \subfloat[VGG-16]{
  	\centering
  	\includegraphics[scale=0.58]{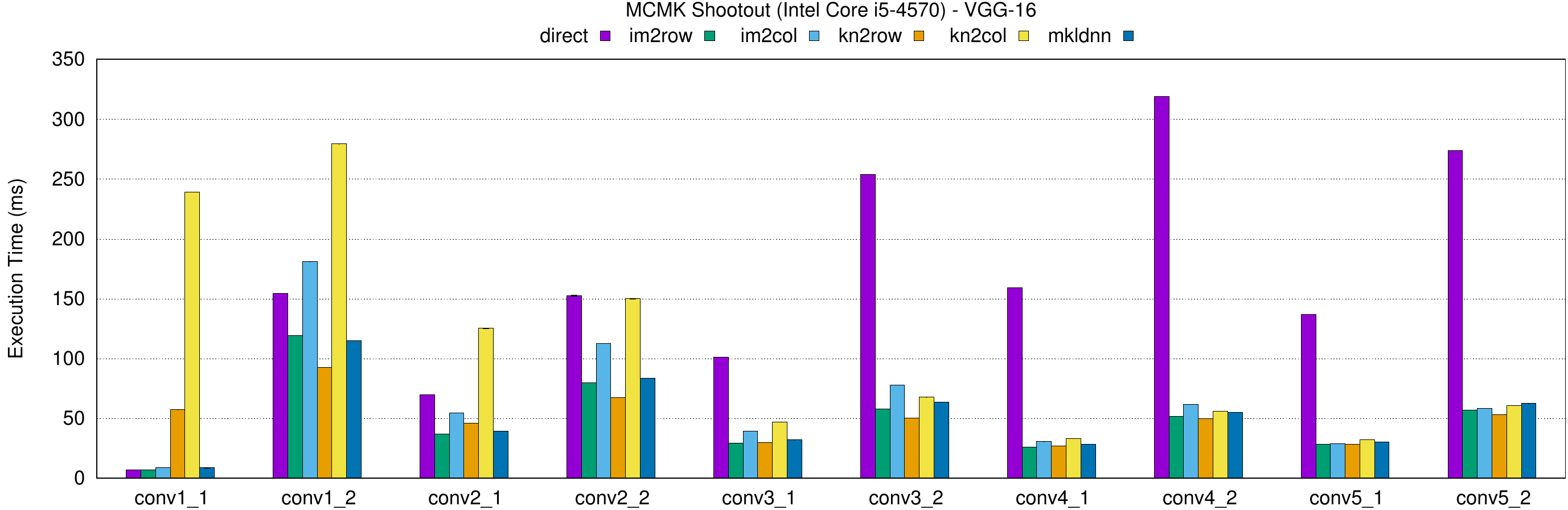}
  	\label{fig:x86-cpu-vgg}
  }\qquad
  \subfloat[Googlenet]{
    \centering
    \includegraphics[scale=0.52]{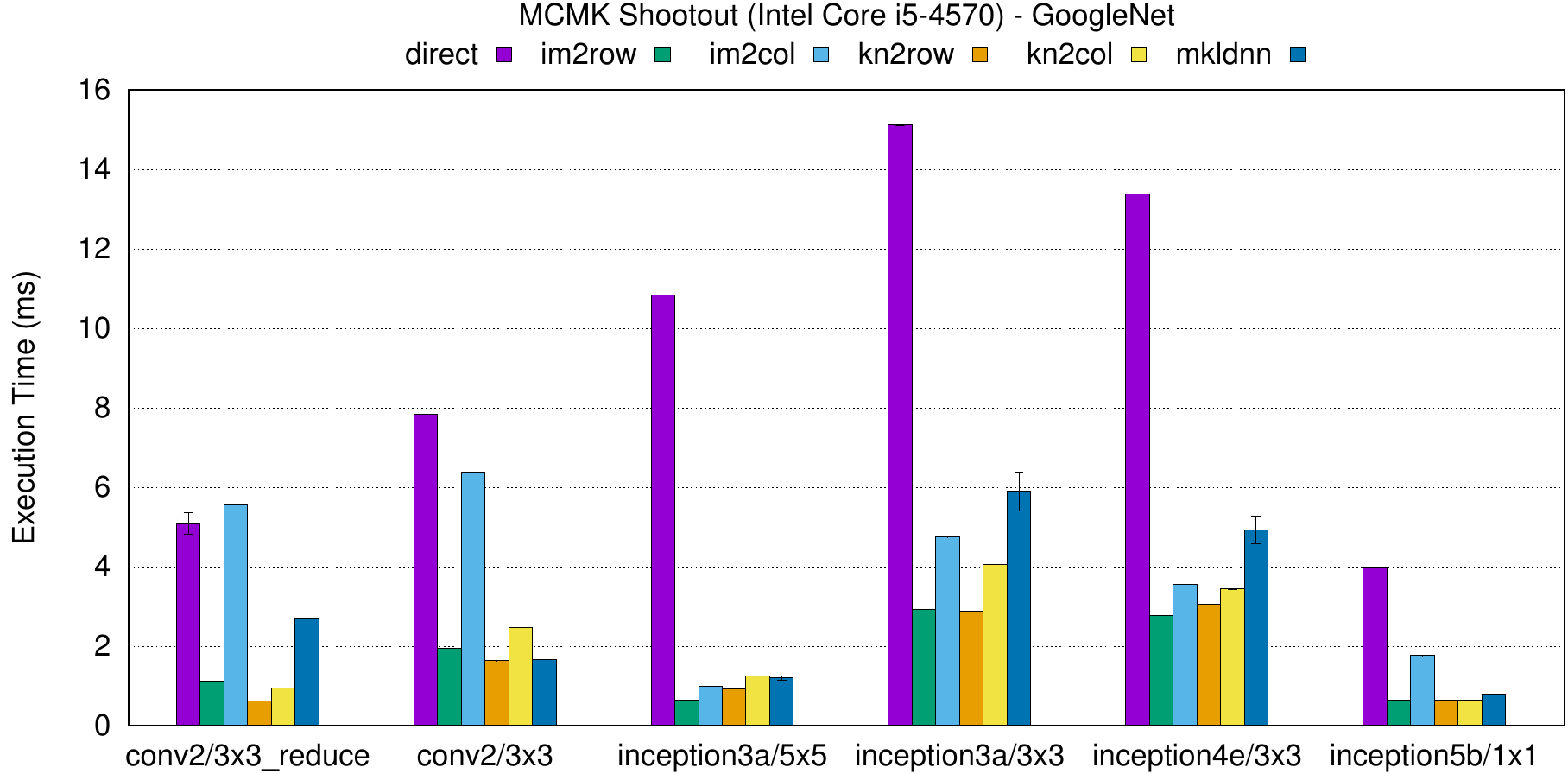}
    \label{fig:x86-cpu-googlenet}
}
  \subfloat[Alexnet]{
    \centering
    \includegraphics[scale=0.52]{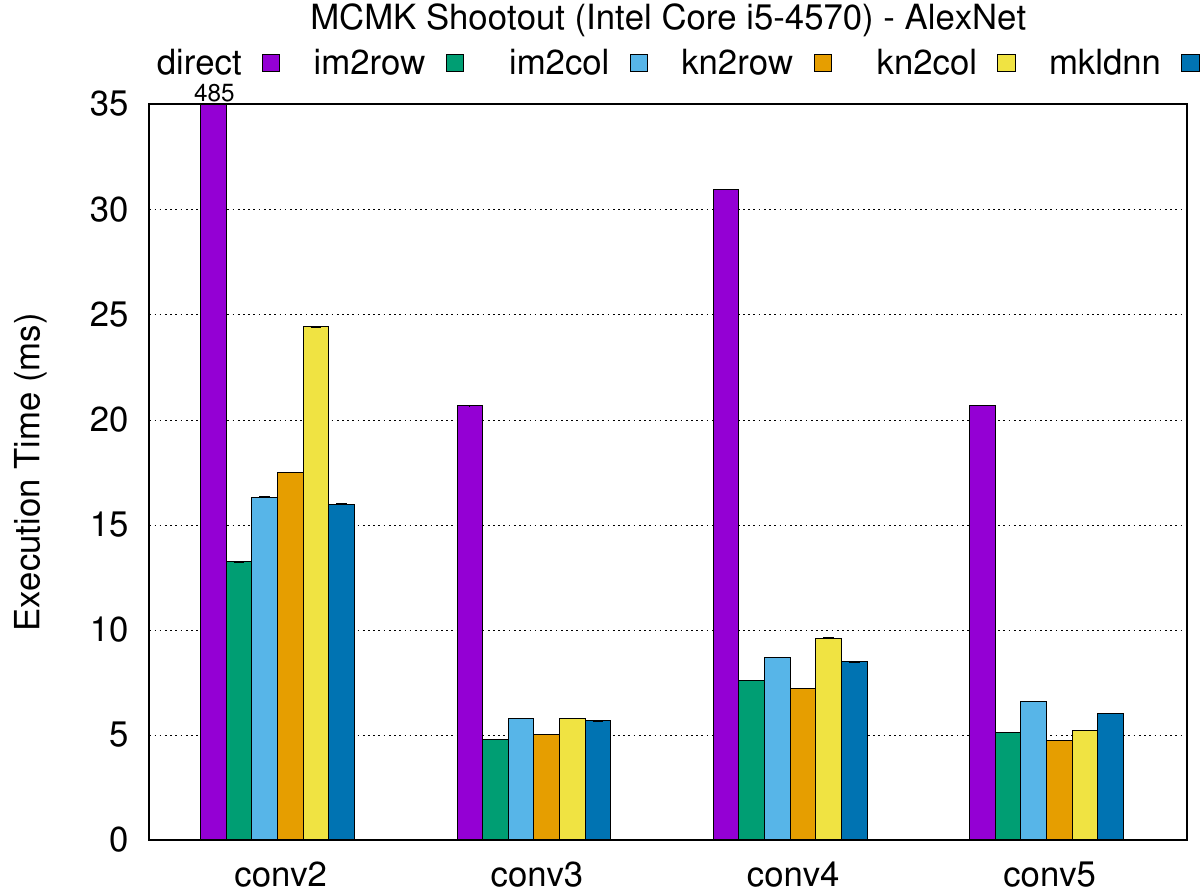}
    \label{fig:x86-cpu-alexnet}
  }

  \caption{Execution time for selected layers of \gls{googlenet}, \gls{vgg} and \gls{alexnet} on the Intel\textregistered ~Core\texttrademark ~i5-4570 CPU. \textbf{Lower is better}. }
  \label{fig:x86-cpu-summary}
\end{figure*}

\subsection{Performance Trends}
\label{sec:perf-tren}

Progressing from left to right across each graph in Figures~\ref{fig:tx1-cpu-summary} and ~\ref{fig:x86-cpu-summary}, the number of input channels increases because the operations are drawn from deeper layers of the \gls{cnn}. At the same time, the size of individual input feature maps diminishes, for the same reason.




Given the fame of \gls{i2c} in the literature, we were surprised to see that the \gls{i2r} method performs so well. When data is laid out in a row matrix instead of a column matrix, spatial locality is significantly improved, since consecutive patch elements are consecutive in memory. While the \gls{i2c} operation may perform well on GPU platforms, our results suggest that it is a poor choice for the implementation of convolution on the CPU.

We also note a large variability between all of the benchmarked methods based on the depth of the convolutional layer in the network. Some methods appear to be very suitable for early layers, but not for later layers; while other methods are unsuitable for early layers, but perform extremely well for later layers. This strongly suggests that a mixture of implementation strategies for convolution is
necessary to achieve peak performance.

For example, direct convolution is very performant for first layer of \gls{vgg}, (Figures~\ref{fig:tx1-cpu-vgg}, ~\ref{fig:x86-cpu-vgg}) but is quickly outpaced by \gls{gemm} based methods as we move deeper in the network. This suggests that peak performance may be achieved by using direct convolution to implement the first layer, and \gls{gemm} based convolution for the remaining layers. However, the situation is different for \gls{alexnet} (Figures~\ref{fig:tx1-cpu-alexnet}, ~\ref{fig:x86-cpu-alexnet}). Here, the \gls{gemm} based methods are always faster.

There is also a similar variability between the \gls{gemm} based methods themselves; some \gls{gemm} based methods are very suitable for early layers, some are very suitable for late layers, but there is no method that has universally good performance in all contexts.


%
%

\setlength\extrarowheight{2pt}

\section{Related Work}
\label{sec:rela-work}

The \gls{i2c} method of performing \gls{mcmk} is an extension of well-known
methods of performing 2D convolution using a Toeplitz matrix. Chellapilla et
al.~\cite{chellapilla2006high} are the first researchers to implement
\gls{mcmk} using using \gls{i2c}. They report significant speedups compared to
the simpler approach of summing multiple channels of 2D convolutions.


Yanqing et al. rediscovered \gls{i2c} for the Caffe deep learning
system~\cite{jia2014caffe}, which uses GPUs and other accelerators to speed up
\glspl{dnn}. The \gls{i2c} approach remains the most widely-used way to
implement \gls{mcmk}, and is used in deep learning frameworks such as Caffe,
Theano and Torch.


Gu et al. \cite{Gu:2016} apply \gls{i2c} to a batch input images to
create a column matrix for multiple input images. They find that
batching can improve throughput by better matching the input matrix
sizes to the optimal sizes for their \gls{gemm} library.


%

Tsai et al. \cite{Tsai:2016} present a set of configurable OpenCL kernels for
\gls{mcmk}. By coding the \gls{mcmk} loop nests directly they eliminate the
need for \gls{i2c} data replication, and thus allow the use of larger batch
sizes while maintaining constraints on local memory. The found that the
performance of a naive loop nest for \gls{mcmk} is not good, but they achieve
satisfactory performance with a program generator and autotuner.

Chetlur et al. \cite{Chetlur:2014} propose a \gls{gemm}-based approach to
convolution based on \gls{i2c}. However, rather than creating the entire column
matrix in one piece, they instead lazily create sub-tiles of the column matrix
in on-chip memory. To optimize performance, they match the size of their
sub-matrix tiles to the tile sizes used by the underlying \gls{gemm}
implementation. They find that this lazy \gls{i2c} achieves speedups over
Caffe's standard \gls{i2c} of between around 0\% and 30\%.



\section{Conclusion} Multi-channel multi-kernel convolution is the most
computationally expensive operation in \glspl{dnn}. Maximal exploitation of
accelerator or processor resources for \gls{mcmk} requires a deep understanding
of the micro-architecture. Careful design of data blocking strategies to
exploit caches, on-chip memories and register locality are needed, along with
careful consideration of data movement and its interaction with SIMD/SIMT
parallelism. Each new processor or accelerator has different performance
characteristics, requiring careful tuning of the code each time it is brought
to a new target.

There are significant advantages in implementing \gls{mcmk} convolution using
existing carefully tuned \glsfirst{gemm} libraries.  However, the most
widely-used approach, \gls{i2c} has a large memory footprint because it
explodes the input image to a much larger column matrix. This space explosion
is quadratic in the radix, $k$ of the convolution being performed. This is
problematic for memory-constrained systems such as embedded object detection
and recognition systems. Additionally, the data redundancy resulting from
\gls{i2c} reduces data locality and increases memory traffic.

We propose new approaches for implementing MCMK convolution using existing
parallel \gls{gemm} libraries. Our \gls{k2r} approach makes one call to
\gls{gemm} and does a post pass on the output to accumulate the partial results
into a single matrix. This drastically increases data locality compared to
\gls{i2c}.



Our results strongly motivate the development of a cost model to guide the
selection of implementations of \gls{mcmk} convolution in deep neural networks. The performance of all of the methods which we evaluate is strongly
context-dependent, with methods having very good performance in some contexts,
and very poor performance in others. The development of this cost model seems a
very fruitful avenue for future work in the area.



\section*{Acknowledgment}

\setlength{\lineskip}{0.5em}
\noindent This work was supported by Science Foundation Ireland
grant 12/IA/1381. This project has received funding from the European Union’s
Horizon 2020 research and innovation programme under grant agreement No 732204
(Bonseyes). This work is supported by the Swiss State Secretariat for
Education‚ Research and Innovation (SERI) under contract number 16.0159. The
opinions expressed and arguments employed herein do not necessarily reflect the
official views of these funding bodies. This work was supported in part by
Science Foundation Ireland grant 13/RC/2094 to Lero --- the Irish Software
Research Centre (www.lero.ie).

%
%

\bibliographystyle{plain}
\bibliography{main}

\end{document}